\documentclass[11pt,a4paper,reqno]{amsart}

\usepackage[margin=27mm,headheight=14pt,headsep=16pt]{geometry}
\usepackage[T1]{fontenc}
\usepackage{lmodern}
\usepackage{microtype}
\allowdisplaybreaks[1]
\usepackage{amsmath,amsthm,amssymb}
\makeatletter
\def\paragraph{\@startsection{paragraph}{4}%
  \z@{1.5ex plus 1ex minus .2ex}%
  {-1em}%
  {\normalfont\normalsize\bfseries}}
\makeatother

\usepackage{enumitem}
\setlist[enumerate]{itemsep=3pt,topsep=5pt,label=\emph{(\alph*)}}
\setlist[enumerate,2]{itemsep=2pt,label=\emph{(\roman*)}}
\usepackage{graphicx}
\usepackage{booktabs}
\usepackage{float}
\usepackage[section]{placeins}
\usepackage{tikz}
\usetikzlibrary{positioning,arrows.meta,fit,backgrounds,calc}
\usepackage[algo2e,ruled,vlined]{algorithm2e}

\usepackage{xcolor}
\definecolor{myblue}{RGB}{0, 61, 201}
\usepackage[numbers,square]{natbib}
\usepackage[hypertexnames=false]{hyperref}
\usepackage{bookmark}
\hypersetup{
  pdftitle={Nonlinear Maps Between Irreducible Representations in Graph Neural Networks},
  pdfauthor={Mani Shayestehfar},
  colorlinks=true,
  linkcolor=myblue,
  citecolor=myblue,
  urlcolor=myblue,
  linktocpage=true
}
\usepackage[nameinlink,noabbrev]{cleveref}
\crefname{section}{Section}{Sections}
\crefname{subsection}{Section}{Sections}
\crefname{appendix}{Appendix}{Appendices}

\newtheorem{theorem}{Theorem}[section]
\newtheorem{corollary}[theorem]{Corollary}
\theoremstyle{definition}
\newtheorem{definition}[theorem]{Definition}
  {\hfill$\square$\par\medskip}

\newcommand{\R}{\mathbb{R}}
\newcommand{\N}{\mathbb{N}}
\newcommand{\E}{\mathbb{E}}
\newcommand{\Sym}{\operatorname{Sym}}
\newcommand{\Hom}{\operatorname{Hom}}

\newcommand{\HomSn}{\operatorname{Hom}_{S_n}}
\newcommand{\He}{\operatorname{He}}
\newcommand{\relu}{\operatorname{ReLU}}

\newcommand{\Specht}[1]{S^{#1}}
\newcommand{\SpechtTriple}{\textnormal{\scshape Specht Triple}}
\newcommand{\Johnson}{\textnormal{\scshape Control Triple}}
\newcommand{\Projector}{\textnormal{\scshape Projector Control}}
\newcommand{\GIN}{\textsc{GIN}}

\newcommand{\GinMSE}{0.09132}
\newcommand{\JohnsonMSE}{0.07640}
\newcommand{\SpechtMSE}{0.06171}

\newcommand{\PrimaryLo}{2.24\%}
\newcommand{\PrimaryHi}{18.27\%}

\newcommand{\ProjectorMSE}{0.06937}
\newcommand{\SpechtGainProjector}{11.05\%}
\newcommand{\DiameterGin}{0.11720}
\newcommand{\DiameterJohnson}{0.10492}
\newcommand{\DiameterSpecht}{0.08645}
\newcommand{\AlgGin}{0.06543}
\newcommand{\AlgJohnson}{0.04787}
\newcommand{\AlgSpecht}{0.03696}

\title[Nonlinear Maps Between Irreducible Representations]{Nonlinear Maps Between Irreducible Representations in Graph Neural Networks}
\author{Mani Shayestehfar}
\address{School of Mathematics and Statistics, The University of Sydney, Sydney, NSW, Australia}
\email{mani.shayestehfar@sydney.edu.au}
\date{}

\urladdr{https://maani.info}

\begin{document}
\begin{abstract}
Equivariant linear maps cannot mix non-isomorphic irreducible representations, whereas nonlinear activations can. We develop a degree-resolved version of this nonlinear interaction picture and use the resulting block inventory to construct \SpechtTriple{}, a residual on unordered graph triples. On a synthetic structural benchmark with a frozen GIN backbone, \SpechtTriple{} reduces aggregate standardised test MSE by \SpechtGainProjector{} relative to a parameter-matched \Projector{} using the same irreducible projectors and complete linear equivariant space. The results support irreducible decomposition as an inductive bias for nonlinear higher-order equivariant processing.
\end{abstract}
\maketitle

\section{Introduction}
\label{sec:intro}

Symmetry is a standard source of inductive bias in machine learning. Convolutional networks exploit translations, graph networks respect relabellings, and many geometric architectures are built so that intermediate features transform equivariantly under a prescribed group symmetry. The geometric deep learning viewpoint organises these constructions by asking which symmetries and geometric structures should be built into the hypothesis class \citep{Bronstein2021GDL}. Representation theory provides the corresponding linear language (The
mathematical concept of representation, which is the only meaning used in this paper, is different from the typical meaning of “representation” in machine learning.).

For linear equivariant maps, the basic picture is especially clean. Decompose a representation, $V$, into irreducible summands, $V = \bigoplus_{i} L_i$, and Schur's lemma restricts every equivariant linear map $L_i \to L_j$ to be nonzero only if $i =j$. However, a neural network is not linear. We consider the bias-free neural architecture
$$A_m \sigma(A_{m-1}\sigma(\cdots \sigma(A_1v)))$$
where $v$ is the input vector, $A_i$ are weight matrices, and $\sigma:\mathbb{R}\to\mathbb{R}$ scalar activation functions. A coordinatewise activation such as ReLU may already be equivariant in the natural permutation basis and yet, after changing to an irreducible basis, it can create nonzero maps between non-isomorphic summands. The nonlinearity therefore carries representation-theoretic information that is invisible to the linear theory.

Gibson, Tubbenhauer and Williamson~\citep{GTW2025} (GTW) proposed studying these nonlinear
equivariant maps under the name \emph{piecewise-linear representation
theory}.  Their starting point is a piecewise-linear
analogue of Schur's lemma: for irreducible, non-isomorphic
$G$-representations $L_i$ and $L_j$,
\begin{equation}\label{eq:pl_schur}
    \Hom_G^{\mathrm{PL}}(L_i,L_j)\neq 0
    \qquad\Longleftrightarrow\qquad
    \ker L_i\subseteq \ker L_j .
\end{equation}
Thus, unlike an equivariant linear map, a piecewise-linear equivariant map
may transfer information between distinct irreducible types.  GTW encode
these possible transfers using \emph{interaction graphs}, whose vertices
are irreducible representations and whose edges record which source and
target types can interact through the nonlinearity. 

Our theoretical contribution refines this existence-level picture by
resolving each possible interaction by degree.  Rather than asking only
whether a nonlinear route $L_i\to L_j$ can occur, we ask at which degrees
$k$ it can occur through
$$
\Hom_G\!\left(\Sym^k L_i,L_j\right).
$$
The resulting factorisation separates two independent ingredients:
representation theory determines which degree-$k$ source-to-target routes
are available, while the spectrum of the activation determines which of those degrees are actually supplied by the chosen nonlinearity. In earlier work \citep{Shayestehfar2026Graded}, we develop the broader theory of these three obstructions; the factorisation below is its projected-block specialisation. The present paper asks whether that structure can guide a learned model. The computational question is:
\begin{quote}
\emph{Can the nonlinear flow between irreducible components predicted by representation theory be exploited as an inductive bias?}
\end{quote}
We study this question through structural graph regression. The targets are graph diameter and algebraic connectivity, two global quantities whose prediction requires information extending beyond a single local neighbourhood. As a node-level baseline we use a Graph Isomorphism Network (GIN) \citep{Xu2019GIN}, which is a message-passing GNN that updates each node by combining its current state with a sum of its neighbours' states through a multilayer perceptron.
After graph-level pooling, the GIN produces a baseline prediction. We freeze this backbone and ask whether a higher-order residual can improve it using features indexed by unordered triples of vertices. Triples are a useful testbed because they carry information beyond ordinary node-level message passing while their $S_n$-representation theory remains explicit: the four irreducible components can be read, roughly, as global, vertex-level, pair-level and genuinely triple-level variation.

This perspective is complementary to the standard GDL programme. GDL and related equivariant architectures use symmetry, often together with irreducible feature types, primarily to constrain the admissible linear maps, while higher-order graph networks extend this principle to features indexed by tuples of vertices \citep{Bronstein2021GDL,Maron2019IGN,Morris2019}. Our question begins one step later. Once equivariance has already been imposed, what structure does a coordinatewise nonlinearity create between the irreducible components of such a representation? Following GTW, we study this nonlinear inter-irrep flow and ask whether exposing it explicitly in a higher-order graph model provides an additional inductive bias.

\paragraph{Contributions.}
The paper makes three contributions.
\begin{enumerate}
\item We present the projected-block degree factorisation, separating symmetry, coordinates and activation (Appendix~\ref{app:gaussian}).
\item We introduce \SpechtTriple{}, a representation-aware higher-order graph residual that resolves unordered triple features into Specht components before nonlinear processing. A worked route explains how the grading constrains the architecture.
\item We compare with two parameter-matched controls. The primary \Projector{} comparison gives a \SpechtGainProjector{} aggregate reduction across ten paired seeds, driven mainly by diameter.
\end{enumerate}

\vspace{0.3cm}
\section{The triple representation}
\label{sec:triple}

Let $S_n$ act on the unordered triples $X_{n,3}:=\binom{[n]}3$. The corresponding permutation module is $V_{n,3}=\R[X_{n,3}]$. For $n\ge6$ it decomposes multiplicity-free as
\begin{equation}
V_{n,3}\cong L_0\oplus L_1\oplus L_2\oplus L_3,
\qquad
L_i=\Specht{(n-i,i)}.
\label{eq:tripledecomp}
\end{equation}
Before using the partition labels, it is useful to read the four summands by the order of interaction they capture.

\begin{table}[H]
\centering
\caption{The irreducible components of the unordered-triple permutation module. The descriptions are an interpretation of the subset-module filtration.}
\label{tab:components}
\small
\begin{tabular}{lll}
\toprule
Interpretation & Specht type & Dimension \\
\midrule
Global variation & $L_0=\Specht{(n)}$ & $1$ \\
Vertex-level variation & $L_1=\Specht{(n-1,1)}$ & $n-1$ \\
Pair-level variation & $L_2=\Specht{(n-2,2)}$ & $n(n-3)/2$ \\
 Triple variation & $L_3=\Specht{(n-3,3)}$ & $n(n-1)(n-5)/6$ \\
\bottomrule
\end{tabular}
\end{table}

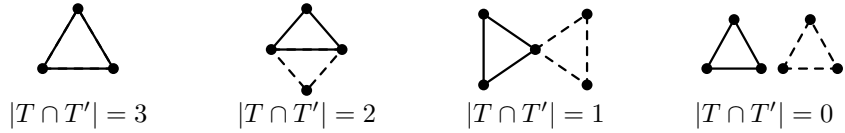
\begin{figure}[H]
\centering

\begin{tikzpicture}[
    scale=0.72,
    line cap=round,
    line join=round,
    triA/.style={draw=black, line width=0.85pt},
    triB/.style={draw=black, line width=0.85pt, dashed},
    vertex/.style={circle, fill=black, inner sep=1.45pt},
    label/.style={font=\small},
]


\begin{scope}[xshift=0cm]
    \path[use as bounding box] (-1.25,-1.55) rectangle (1.25,1.05);

    \coordinate (a) at (-0.65,-0.35);
    \coordinate (b) at ( 0.65,-0.35);
    \coordinate (c) at ( 0.00, 0.75);

    \draw[triA] (a)--(b)--(c)--cycle;
    \draw[triB] (a)--(b)--(c)--cycle;

    \foreach \p in {a,b,c}
        \node[vertex] at (\p) {};

    \node[label] at (0,-1.20) {$|T\cap T'|=3$};
\end{scope}

\begin{scope}[xshift=4.2cm]
    \path[use as bounding box] (-1.25,-1.55) rectangle (1.25,1.05);

    \coordinate (a) at (-0.65,0);
    \coordinate (b) at ( 0.65,0);
    \coordinate (c) at ( 0.00,0.75);
    \coordinate (d) at ( 0.00,-0.75);

    \draw[triA] (a)--(b)--(c)--cycle;
    \draw[triB] (a)--(b)--(d)--cycle;

    \foreach \p in {a,b,c,d}
        \node[vertex] at (\p) {};

    \node[label] at (0,-1.20) {$|T\cap T'|=2$};
\end{scope}

\begin{scope}[xshift=8.4cm]
    \path[use as bounding box] (-1.25,-1.55) rectangle (1.25,1.05);

    \coordinate (a) at (0,0);
    \coordinate (b) at (-0.95, 0.65);
    \coordinate (c) at (-0.95,-0.65);
    \coordinate (d) at ( 0.95, 0.65);
    \coordinate (e) at ( 0.95,-0.65);

    \draw[triA] (a)--(b)--(c)--cycle;
    \draw[triB] (a)--(d)--(e)--cycle;

    \foreach \p in {a,b,c,d,e}
        \node[vertex] at (\p) {};

    \node[label] at (0,-1.20) {$|T\cap T'|=1$};
\end{scope}

\begin{scope}[xshift=12.6cm]
    \path[use as bounding box] (-1.25,-1.55) rectangle (1.25,1.05);

    \coordinate (a) at (-1.05,-0.35);
    \coordinate (b) at (-0.05,-0.35);
    \coordinate (c) at (-0.55, 0.55);

    \coordinate (d) at ( 0.35,-0.35);
    \coordinate (e) at ( 1.35,-0.35);
    \coordinate (f) at ( 0.85, 0.55);

    \draw[triA] (a)--(b)--(c)--cycle;
    \draw[triB] (d)--(e)--(f)--cycle;

    \foreach \p in {a,b,c,d,e,f}
        \node[vertex] at (\p) {};

    \node[label] at (0,-1.20) {$|T\cap T'|=0$};
\end{scope}

\end{tikzpicture}

\caption{The four $S_n$-orbits on ordered pairs $(T,T')$ of unordered
triples, classified by their intersection size $|T\cap T'|$.
These relations generate the Johnson scheme $J(n,3)$ and hence the
complete linear $S_n$-equivariant algebra on $V_{n,3}$.}
\label{fig:intersections}
\end{figure}

Two unordered triples are distinguished, up to $S_n$, only by the size of their intersection. Figure~\ref{fig:intersections} shows the four possibilities. Their relation matrices span the Johnson Bose--Mesner algebra. Its primitive orthogonal idempotents $E_0,\ldots,E_3$ are exactly the projectors onto the four summands in \eqref{eq:tripledecomp}. Operationally, $E_0$ extracts global variation, $E_1$ variation attributable to individual vertices, $E_2$ residual pair-level variation, and $E_3$ the remaining genuinely triple-level variation. The closed formulas and their derivation are given in Appendix~\ref{app:johnson}.

\vspace{0.3cm}
\section{What the activation adds}
\label{sec:theory}

Let $H=\R^{\mathcal B}=\bigoplus_i L_i$ be a multiplicity-free permutation representation, with orthogonal projections $p_i:H\to L_i$, inclusions $\iota_i:L_i\hookrightarrow H$, and projector matrices $E_i=\iota_ip_i$. Write $\sigma_{\mathcal B}$ for the coordinatewise extension of a scalar activation $\sigma$ in the permutation basis $\mathcal B$. The nonlinear interaction from $L_i$ to $L_j$ is
\begin{equation}
F_{j\leftarrow i}=p_j\circ\sigma_{\mathcal B}\circ\iota_i:L_i\longrightarrow L_j.
\label{eq:block}
\end{equation}
Unlike the linear case, nonlinear activations can transfer information between non-isomorphic irreducible components. To describe these interactions by degree, we place $F_{j\leftarrow i}$ in a Gaussian $L^2$ space, where it admits an orthogonal Hermite expansion. The degree-$k$ component is then governed by the symmetric power $\Sym^kL_i$, and possible transfers to $L_j$ are encoded by $\Hom_G(\Sym^kL_i,L_j)$.
The activation can retain only those degrees whose coordinate maps are nonzero. A nonzero Hom space is necessary but is not sufficient for this particular block. We denote the degree-$k$ Gaussian-reference component by $\Pi_kF_{j\leftarrow i}$. The formal construction is given in Appendix~\ref{app:gaussian}.

\begin{theorem}[Degree-graded activation block]
\label{thm:factor}
For a continuous scalar activation $\sigma$ with finitely many affine pieces, applied coordinatewise in a transitive permutation basis, let $F_{j\leftarrow i}$ be as in \eqref{eq:block}. Then
\begin{equation}
\Pi_kF_{j\leftarrow i}
=
\frac{a_k(\sigma;s_i)}{s_i^k}\,\Pi_kQ_k^{ji}.
\label{eq:factor}
\end{equation}
Here $s_i^2=\dim L_i/\dim H$ is the coordinate variance for a standard Gaussian on $L_i$, $a_k$ is defined in \eqref{eq:ak-app}, and $Q_k^{ji}(x)=p_j((\iota_i x)^{\odot k})$ is the homogeneous coordinate polynomial. Homogeneous degree-$k$ equivariant polynomials are identified with $\Hom_G(\Sym^kL_i,L_j)$, and $\Pi_kQ_k^{ji}$ is the Hermite representative of $Q_k^{ji}$. The notation $(\iota_i x)^{\odot k}$ means coordinatewise power in the permutation basis. The map depends on those coordinates, not on the activation, and may be zero. Appendix~\ref{app:gaussian} proves the factorisation.
\end{theorem}

\begin{corollary}[Odd-degree blindness]
\label{cor:odd-blindness}
For a scalar activation $\sigma$ homogeneous of degree one, including ReLU,
$$
a_k(\sigma;s_i)=0
\qquad
\text{for every odd }k>1.
$$
Hence $\Pi_kF_{j\leftarrow i}=0$ for every odd $k>1$. This is a statement about a bias-free block; it does not impose the same restriction on arbitrary compositions or on biased activations.
\end{corollary}

Note that the implemented architecture does not explicitly separate Hermite degrees. We use the grading only to determine which complete source-to-target blocks can be nonzero for ReLU.

\paragraph{A worked example.}
Consider the permutation representation of $S_3$,
$$
\mathbb R^3
=
\underbrace{\operatorname{span}\{(1,1,1)\}}_{L_0}
\oplus
\underbrace{\{x\in\mathbb R^3:x_1+x_2+x_3=0\}}_{L_1}.$$
Schur's lemma gives
$\Hom_{S_3}(L_1,L_0)=0$, so there is no linear equivariant route from $L_1$ to $L_0$. Theorem~\ref{thm:factor} asks whether such a route can appear at degree two. Since
$$
\dim\Hom_{S_3}(\Sym^2L_1,L_0)=1,
$$
the homogeneous coordinate map is nonzero: $Q_2^{01}(x)=p_0((\iota_1x)^{\odot2})=\|x\|^2\mathbf1/3$, where $\mathbf1=(1,1,1)$. Since $s_1^2=2/3$, the normalisation in Theorem~\ref{thm:factor} gives
\[
\Pi_2Q_2^{01}(x)=\frac13(\|x\|^2-2)\mathbf1,
\qquad a_2(\relu;s_1)=\frac{s_1}{2\sqrt{2\pi}}>0.
\]
The centring constant is $\dim L_1=2$. Thus $\Pi_2F_{0\leftarrow1}=[a_2(\relu;s_1)/s_1^2]\Pi_2Q_2^{01}\ne0$: ReLU produces a second-chaos interaction where the linear Hom space vanishes. For the triple representation used below, the same calculation is performed for the four Specht components. The resulting route inventory is given in Appendix~\ref{app:routes}.

\vspace{0.3cm}
\section{Representation-aware triple processing}
\label{sec:architecture}

We now use these blocks to define a residual architecture. A five-layer GIN produces node states and a graph-level baseline prediction. The GIN is frozen before any higher-order residual is trained. All three residuals receive the same form of symmetric triple lift, and the final prediction is the frozen GIN output plus the learned residual. For every benchmark order $15\le n\le24$, exact evaluation of the coordinate maps gives the ReLU route set
$$
\mathcal{R} = \{(0,0)\} \cup \{(i,j) \::\: i \in \{1,2,3\}, \: j \in \{0,1,2,3\}\}.
$$
Thus $L_0$ can only map to $L_0$, whereas each nontrivial source has routes to all four targets. All these routes already occur in ReLU-supported degrees at most two, except $L_1 \to L_3$, whose first ReLU-supported contribution occurs in degree four.

\paragraph{Triple lift.}
Let $\mathcal G=(V,E)$ be a simple graph with $|V|=n$, and let
$A$ denote its adjacency matrix,
$$
A_{uv}
=
\begin{cases}
1, & \{u,v\}\in E,\\
0, & \text{otherwise},
\end{cases}
\qquad A_{uu}=0.
$$
The frozen GIN assigns to each vertex $v\in V$ a hidden state
$h_v\in\R^{64}$, where $64$ is the hidden width of the backbone.
To apply the $S_n$-representation theory of unordered triples, we convert
these node-level states into a feature indexed by every
$T\in\binom V3$.

For a triple $T=\{u,v,w\}$, we retain two kinds of information. First, the
three frozen node states are projected to the channel space of the
higher-order branch by a learned map $W_h:\R^{64}\to\R^6$ and then averaged.
Second, we record the connectivity internal to the triple. Since there are
exactly three possible edges among $u,v,w$, define
$$
c(T)
:=
A_{uv}+A_{uw}+A_{vw}
\in\{0,1,2,3\}.
$$
Thus $c(T)$ is simply the number of edges in the subgraph induced by $T$.
A learned embedding
$e:\{0,1,2,3\}\to\R^6$ converts this discrete structural quantity into the
same six-dimensional channel space. We then define
\begin{equation}
X_T
=
\frac{1}{3}
\bigl(W_hh_u+W_hh_v+W_hh_w\bigr)
+
e\!\left(c(T)\right)
\in\R^6.
\label{eq:lift}
\end{equation}
The dimension $6$ is an architectural choice for the width of the
higher-order residual branch; it is not prescribed by the representation
theory.

Collecting the features of all $N=\binom n3$ unordered triples gives
$$
X=(X_T)_{T\in\binom V3}
\in
V_{n,3}\otimes\R^6
\cong
\R^{N\times6}.
$$
Both terms in \eqref{eq:lift} depend only on the unordered set $T$.
Consequently, a vertex relabelling $\pi\in S_n$ sends the coordinate
$X_T$ to $X_{\pi(T)}$. The triple lift is therefore $S_n$-equivariant,
with $S_n$ acting on the $N$ triple coordinates and trivially on the six
feature channels.

\paragraph{Specht Triple.}
The exact Johnson projectors give $X_i=E_iX$. If $d_i=\dim L_i$, the source is rescaled by $\sqrt{N/d_i}$, matching its Gaussian-reference unit scale, and multiplied by a learned positive channel gain. ReLU is applied in the original triple-coordinate basis, after which each source activation is projected onto its permitted target components. We write
\begin{equation}
Y_{j\leftarrow i}
=E_j\relu\!\left(g_i\odot\sqrt{N/d_i}\,E_iX\right)
\in\R^{N\times6},
\label{eq:route}
\end{equation}
where $i$ is the source and $j$ the target. For a route tensor
$Y=(Y_T)_{T\in\binom V3}\in\R^{N\times6}$,
we remove the triple index using two permutation-invariant channelwise
statistics, defined as
\begin{equation}
\Phi(Y)=\left(\frac1N\sum_TY_T^2,\;\frac1N\sum_T\relu(Y_T)\right)\in\R^{12}.
\label{eq:pool}
\end{equation}
The projectors act as $E_i\otimes I_6$, and the source gains and ReLU act channelwise. Each channel therefore uses the multiplicity-free block theorem separately (Appendix~\ref{app:gaussian}). A learned gate acts on each route summary before accumulation. Algorithm~\ref{alg:specht} states the implementation with shapes.

\begin{algorithm2e}[H]
\caption{\SpechtTriple{} residual for one graph.}
\label{alg:specht}
\DontPrintSemicolon
\SetAlgoLined
\KwIn{$A\in\{0,1\}^{n\times n}$, frozen $H_{\mathrm{GIN}}\in\R^{n\times64}$ and $\hat y_0\in\R^2$; projectors $E_i\in\R^{N\times N}$; route set $\mathcal R$.}
\KwOut{Prediction $\hat y\in\R^2$.}
$X\leftarrow\operatorname{TripleLift}(H_{\mathrm{GIN}},A)\in\R^{N\times6}$ using \eqref{eq:lift}\;
\For{$i=0,1,2,3$}{
  $X_i\leftarrow E_iX\in\R^{N\times6}$\;
  $V_i\leftarrow\relu(g_i\odot\sqrt{N/d_i}\,X_i)\in\R^{N\times6}$\;
}
$z\leftarrow0\in\R^{12}$\;
\ForEach{$(i,j)\in\mathcal R$}{
  $Y_{j\leftarrow i}\leftarrow E_jV_i\in\R^{N\times6}$\;
  $z\leftarrow z+\gamma_{j\leftarrow i}\odot\Phi(Y_{j\leftarrow i})$\;
}
$r\leftarrow\operatorname{Head}(\operatorname{LayerNorm}(z/\sqrt{48}))\in\R^2$\;
\Return{$\hat y_0+r$}
\end{algorithm2e}

\paragraph{Control Triple.}
To test whether the gain of \SpechtTriple{} comes from its irreducible
organisation of the nonlinearity, we compare it with a control that already
has access to the full space of linear $S_n$-equivariant mixing between
unordered triples.

Let $N=\binom n3$ and let $A_1\in\R^{N\times N}$ be the adjacency matrix of
the Johnson graph, where
$$
(A_1)_{T,T'}=1
\qquad\Longleftrightarrow\qquad
|T\cap T'|=2.
$$
Since every triple has $3(n-3)$ such neighbours, define the normalised Johnson operator $T_n:=\frac{1}{3(n-3)}A_1.$
From its lifted tensor $X\in\R^{N\times6}$ defined by \eqref{eq:lift}, the
control computes
\begin{equation}
L_J(X)
=
\sum_{r=0}^{3}T_n^r XW_r,
\qquad
W_r\in\R^{6\times6}.
\label{eq:johnsoncontrol}
\end{equation}
Here $T_n^r$ mixes triple coordinates and $W_r$ mixes feature channels. At
fixed $n$, the four operators $I,T_n,T_n^2,T_n^3$ span
$\text{End}_{S_n}(V_{n,3})$, so \Johnson{} can realise arbitrary linear
$S_n$-equivariant mixing before the activation.

The distinction between the two residuals is where the nonlinearity is applied. Control Triple first forms a general linear $S_n$-equivariant mixture and then applies ReLU. Specht Triple first separates the four irreducible source components, applies ReLU to each source separately, projects the result onto the permitted targets in $\mathcal{R}$, and pools the resulting route summaries. Further Johnson-scheme details and
parameter accounting are given in
Appendices~\ref{app:johnson} and~\ref{app:protocol}.

\paragraph{\Projector{}.}
The stronger control uses the same primitive projectors as \SpechtTriple{}, forming $L_P(X)=\sum_{i=0}^3E_iXU_i$ with learned $U_i\in\R^{6\times6}$ before one ReLU. Both controls apply $\relu(L(X)/\sqrt4+b)$, then \eqref{eq:pool}, layer normalisation and a residual head. At fixed $n$, $\{E_i\}_{i=0}^3$ and $\{T_n^r\}_{r=0}^3$ span the same complete linear equivariant algebra. \Projector{} therefore controls for access to the irreducible projectors as well as the linear space. It does not retain nonlinear source-to-target blocks separately.

\vspace{0.3cm}
\section{Experimental design}
\label{sec:experiment}

The frozen \GIN{} provides node-level message passing and a baseline graph prediction.  \Johnson{} adds global linear $S_n$-equivariant mixing of triple features, \Projector{} uses the same linear space in the projector basis, and
\SpechtTriple{} uses the same triple lift but resolves the nonlinear computation by irreducible source and target type. Each residual has $884$ trainable parameters. Equal parameter counts do not equalise nonlinear factorisation, normalisation or computation, so this is a comparison of complete residual designs rather than an isolation of one mechanism. Hence, the primary comparison is \SpechtTriple{} against \Projector{}.

We use a synthetic structural benchmark so that the targets have clear
graph-theoretic meaning and the graph orders remain small enough for exact
triple representations. Graphs have $15\le n\le24$ and are sampled from
Erd\H{o}s--R\'enyi, Barab\'asi--Albert, Watts--Strogatz and two-block
stochastic block families. The primary regression targets are diameter,
which measures a global shortest-path scale, and algebraic connectivity,
the second-smallest eigenvalue of the combinatorial Laplacian. These probe
different aspects of global graph structure while using the same inputs.

Exact-$n$ batching ensures that the Specht projectors and Johnson operators
always correspond to the actual $S_n$ action of each graph. The complete
data split, generator parameters, architecture dimensions, optimisation
settings, random-identifier protocol, ten paired seeds and statistical
procedure are reported in Appendix~\ref{app:protocol};
Table~\ref{tab:setup} gives a compact summary.

\vspace{0.3cm}
\section{Results}
\label{sec:results}

The comparison of the three models is summarised in Table~\ref{tab:results}. Across ten paired seeds, \SpechtTriple{} improves on \Johnson{} in every seed. It reduces aggregate standardised test MSE by $19.23\%$ with paired-bootstrap interval $[15.97\%, 21.90\%]$. Against \Projector{}, the reduction is \SpechtGainProjector{} with paired-bootstrap 95\% interval $[\PrimaryLo{},\PrimaryHi{}]$ and $7/10$ seed wins.

\SpechtTriple{} improves on \Johnson{} for both diameter and algebraic connectivity. Importantly, diameter improves by $14.24\%$ with $8/10$ seed wins, whereas algebraic connectivity improves by $2.57\%$ with $6/10$ wins. Hence the strongest empirical conclusion is a benefit for diameter on this benchmark. Complete per-seed values, paired uncertainty estimates, hypothesis tests, taskwise reductions, graph-family breakdowns and graph-order diagnostics are reported in Appendix~\ref{app:protocol}.

\begin{table}[H]
\centering
\caption{Mean standardised test MSE over ten paired seeds. The primary comparison is \SpechtTriple{} versus \Projector{}. All residuals have $884$ trainable parameters.}
\label{tab:results}
\small
\begin{tabular}{lrrrr}
\toprule
Model & Residual params & Aggregate & Diameter & Alg. connectivity \\
\midrule
\GIN{} & -- & \GinMSE{} & \DiameterGin{} & \AlgGin{} \\
\Johnson{} & 884 & \JohnsonMSE{} & \DiameterJohnson{} & \AlgJohnson{} \\
\Projector{} & 884 & \ProjectorMSE{} & 0.10081 & 0.03794 \\
\SpechtTriple{} & 884 & \textbf{\SpechtMSE{}} & \textbf{\DiameterSpecht{}} & \textbf{\AlgSpecht{}} \\
\bottomrule
\end{tabular}
\end{table}

\vspace{0.3cm}
\section{Discussion}
\label{sec:discussion}

The experiment supports a stronger claim than the earlier pointwise-triple control could justify. \Johnson{} can already mix every triple with every other triple through the full linear $S_n$-equivariant Johnson algebra before applying ReLU. Its improvement over GIN confirms that higher-order equivariant processing is useful on the benchmark. The further \SpechtTriple{} gain therefore points to the organisation of nonlinear computation by irreducible source and target type, rather than simply to adding triples or allowing global triple mixing.

The comparison should not be read as a pure change-of-basis experiment. \SpechtTriple{} and \Johnson{} have different nonlinear factorisations. The Johnson branch applies one general equivariant linear map and then ReLU. The Specht branch resolves sources, applies ReLU separately, projects onto targets, and preserves the resulting block summaries until the invariant readout. Representation-adapted coordinates may also provide a better conditioned parameterisation of the equivariant algebra. These effects are all part of the representation-aware inductive bias tested here, so the $19.23\%$ gap should not be attributed causally to one individual route mechanism.

The theory explains why this design is natural. Linear equivariance alone sees the four Specht types as decoupled channels. GTW show that the activation reconnects them. The degree grading identifies when those reconnections can occur and which ones a particular activation can supply. \SpechtTriple{} does not explicitly separate Hermite degrees, but it uses the grading to determine and audit the ReLU block inventory, then exposes those full nonlinear blocks to learning. An explicit degree-resolved architecture is a separate direction.

\paragraph{Limitations.}
The GIN backbone is frozen, so the experiment does not show that \SpechtTriple{} outperforms end-to-end higher-order graph architectures. The Gaussian construction provides a reference grading for the nonlinear maps, but it does not assume that the learned GIN features themselves are Gaussian. Although \Johnson{} spans the full linear equivariant algebra at each fixed graph order, the same polynomial-basis parameters are shared across different values of $n$. It therefore does not learn an independent Johnson operator for each graph size. The random node identifiers preserve permutation invariance when they are relabelled together with the vertices, but predictions are not invariant to drawing a new set of identifiers. Finally, the experiments cover only one family of groups, one $k$-subset representation and one structural graph benchmark.

\begin{samepage}
\paragraph{Conclusion.}
For equivariant linear maps, irreducible decomposition simplifies the problem. For nonlinear equivariant maps, the activation creates structured communication between those irreducibles. This paper shows that the same structure can be used computationally. Generic equivariant triple mixing already improves a message-passing backbone, while resolving the nonlinear computation by Specht source and target gives further improvements. The key takeaway should be that piecewise-linear representation theory is not only a language for interpreting equivariant networks, but it also is a source of architectural inductive bias.
\end{samepage}

\vspace{0.3cm}
\paragraph{Code availability.}
The code used to generate the experiments in this paper, including data generation, all three residual models, training scripts, evaluation routines and representation-theoretic operators, is available in this \href{https://github.com/ManiShayestehfar/NeurReps-Codebase-Addendum}{GitHub repository}.

\vspace{0.3cm}
\section*{Acknowledgements}
I would like to thank Dani Tubbenhauer for their guidance and reviews. I also thank the anoynomous reviewers of \href{https://neurreps.org/}{NeurReps 2026 Workshop} for their helpful comments.

\vspace{0.3cm}
\section*{AI Declaration}
OpenAI's ChatGPT 5.6 was used initially for development of the polynomial--Hermite decomposition theory, followed by assistance with computational experiment design, code generation and debugging, computational analysis, and the editing and restructuring of the paper. The author takes full responsibility for the mathematical statements, experimental design, reported results, citations, and final text.

\clearpage
\bibliographystyle{alpha}
{\raggedright
\def\bibliofont{\small}
\bibliography{references}
}

\clearpage
\appendix

\section{Representation theory and the Johnson scheme}
\label{app:johnson}

This appendix collects the linear representation theory used by the main text. A permutation representation is a space $\R^{\mathcal B}$ with an orthonormal basis permuted by $G$. The action is \emph{transitive} if any basis vector can be sent to any other by a group element. It is \emph{multiplicity-free} if its irreducible summands are pairwise non-isomorphic. These are distinct conditions: transitivity equates coordinate variances, while multiplicity-freeness labels the blocks without choices among repeated copies. Let $G$ be finite and let all representations be finite-dimensional, real and orthogonal. A linear map $\varphi:L\to K$ is $G$-equivariant when $\varphi(gx)=g\varphi(x)$. If $L$ and $K$ are irreducible and nonisomorphic, Schur's lemma gives $\Hom_G(L,K)=0$. Thus an equivariant linear map is block diagonal across non-isomorphic irreducible types in a multiplicity-free decomposition.

For the action of $S_n$ on $r$-subsets, the permutation module $V_{n,r}$ is the Young permutation module $M^{(n-r,r)}$. For $r=3$ and $n\ge6$, the classical Young rule gives
$$
V_{n,3}\cong\Specht{(n)}\oplus\Specht{(n-1,1)}\oplus\Specht{(n-2,2)}\oplus\Specht{(n-3,3)},
$$
with dimensions $\binom ni-\binom n{i-1}$ for $i=0,1,2,3$ \citep{JamesKerber1981}.

\subsection{Johnson relations and exact projectors}

Let $A_t$ be the matrix whose $(T,T')$ entry is one when $|T\cap T'|=3-t$. The four matrices $A_0=I,A_1,A_2,A_3$ span the Bose--Mesner algebra of the Johnson scheme $J(n,3)$ \citep{Delsarte1973,BCN1989}. Since the triple module is multiplicity-free, this algebra is exactly $\operatorname{End}_{S_n}(V_{n,3})$.

The eigenvalue of $A_1$ on $L_i=\Specht{(n-i,i)}$ is
$$
\theta_i=(3-i)(n-3-i)-i,
\qquad i=0,1,2,3.
$$
These values are pairwise distinct for $n\ge6$. Consequently the primitive idempotent projecting onto $L_i$ is
\begin{equation}
E_i=\prod_{j\ne i}\frac{A_1-\theta_jI}{\theta_i-\theta_j}.
\label{eq:projector-app}
\end{equation}
This gives the exact-$n$ projectors used by \SpechtTriple{}. The implementation applies the polynomial in $A_1$ sparsely rather than constructing a dense $N\times N$ matrix.

\subsection{Why both controls are linearly complete}

Set $T_n=A_1/[3(n-3)]$. Normalisation changes the eigenvalues but not their distinctness. Since $T_n$ has four distinct eigenvalues, its minimal polynomial has degree four. Therefore
$$
\operatorname{span}\{I,T_n,T_n^2,T_n^3\}=\operatorname{End}_{S_n}(V_{n,3}).
$$
For six feature channels on which $S_n$ acts trivially,
$$
\operatorname{End}_{S_n}(V_{n,3}\otimes\R^6)
\cong
\operatorname{End}_{S_n}(V_{n,3})\otimes\operatorname{End}(\R^6).
$$
Hence the layer $\sum_{r=0}^3T_n^rXW_r$ with arbitrary $W_r\in\R^{6\times6}$ parameterises the complete linear equivariant operator space at fixed $n$. The projector-basis form $\sum_iE_iXU_i$ parameterises the same space. Each mixing layer has $4\cdot6^2=144$ weights, exactly the dimension predicted by the commutant calculation. This fixed-$n$ spanning statement does not equate their parameterisations when weights are shared across different graph orders.

\vspace{0.3cm}
\section{Gaussian/Hermite background and proofs}
\label{app:gaussian}

The degree grading is an exact decomposition of the nonlinear map with respect to a Gaussian reference measure. This section defines the relevant spaces before proving Theorem~\ref{thm:factor}.

\subsection{Gaussian spaces}

\begin{definition}[Standard Gaussian measure]
Let $L$ be an $m$-dimensional real inner-product space. The standard Gaussian measure $\gamma_L$ is the probability measure whose density in any orthonormal coordinates is
$$
d\gamma_L(x)=(2\pi)^{-m/2}e^{-\|x\|^2/2}\,dx.
$$
It is invariant under the orthogonal group $\mathrm O(L)$.
\end{definition}

\begin{definition}[Vector-valued Gaussian space]
For a finite-dimensional inner-product space $K$, let $L^2(\gamma_L;K)$ be the Hilbert space of measurable maps $F:L\to K$, modulo equality almost everywhere, with
$$
\|F\|_{L^2(\gamma_L;K)}^2=\int_L\|F(x)\|_K^2\,d\gamma_L(x)<\infty.
$$
\end{definition}

A continuous piecewise-linear map with finitely many pieces has at most linear growth, hence belongs to this space. If $L$ and $K$ are orthogonal $G$-representations, $G$ acts unitarily on $L^2(\gamma_L;K)$ by
$$
(g\cdot F)(x)=g\,F(g^{-1}x),
$$
and the fixed points are precisely the square-integrable $G$-equivariant maps.

\subsection{Hermite polynomials and polynomial chaos}

We use the \emph{probabilists' Hermite polynomials}, defined by
\[
e^{tz-t^2/2}=\sum_{k\ge0}\He_k(z)\frac{t^k}{k!}.
\]
Thus $\He_0=1$, $\He_1=z$, $\He_2=z^2-1$, $\He_3=z^3-3z$, and $\E[\He_j(Z)\He_k(Z)]=k!\delta_{jk}$ for $Z\sim\mathcal N(0,1)$. They are orthogonal rather than unit-normalised, and complete in $L^2(\gamma_\R)$ \citep{Janson1997}.

In orthonormal coordinates on $L\cong\R^m$, the products $\prod_{r=1}^m\He_{\alpha_r}(x_r)$ with $\sum_r\alpha_r=k$ span the $k$th polynomial chaos $\mathcal H_k(L)$. The orthogonal projection onto this space is $\Pi_k$. Orthogonal changes of coordinates preserve each chaos, and the map taking a homogeneous degree-$k$ polynomial to its $k$th-chaos projection is an equivariant isomorphism. In particular, chaos polynomials need not be homogeneous: $x^2-1$ has degree two and a constant term, but is orthogonal to all lower degrees.

Using the invariant inner product to identify $L$ with its dual gives
\begin{equation}
L^2(\gamma_L;K)^G\cong\widehat{\bigoplus}_{k\ge0}\Hom_G(\Sym^kL,K).
\label{eq:equivchaos-app}
\end{equation}
Indeed, tensor the scalar chaos decomposition with $K$ and take $G$-fixed vectors; each orthogonal chaos projection commutes with the group action. The completion is a Hilbert direct sum. This is an ambient decomposition, not an assertion that every element is piecewise linear. See Shayestehfar~\citep[Section~1.3 and Theorem~2.1]{Shayestehfar2026Graded}.

\subsection{Coordinatewise factorisation}

Let $(H,\mathcal B)$ be a transitive permutation representation with multiplicity-free decomposition $H=\bigoplus_iL_i$, projections $p_i$, inclusions $\iota_i$, and projector matrices $E_i=\iota_ip_i$. For $b\in\mathcal B$, set $a_b=\iota_i^*b=p_ib\in L_i$. Transitivity and equivariance imply that $s_i=\|a_b\|=\|E_ib\|$ is independent of $b$. Since $E_i$ is an orthogonal projector, $s_i^2=(E_i)_{bb}$ and summing over the $\dim H$ basis vectors gives $s_i^2=\dim L_i/\dim H>0$. For $x\in L_i$ define $\xi_b(x)=\langle x,a_b\rangle/s_i$, which is standard Gaussian under $\gamma_{L_i}$. For an activation $\sigma$ define
\begin{equation}
a_k(\sigma;s)=\frac1{k!}\E[\sigma(sZ)\He_k(Z)].
\label{eq:ak-app}
\end{equation}
The activation block is
$$
F_{j\leftarrow i}(x)=\sum_{b\in\mathcal B}\sigma(\langle x,a_b\rangle)p_jb.
$$
Expanding the scalar activation in Hermite polynomials gives
$$
F_{j\leftarrow i}(x)
=\sum_{k\ge0}a_k(\sigma;s_i)
\underbrace{\sum_{b\in\mathcal B}\He_k(\xi_b(x))p_jb}_{s_i^{-k}\Pi_kQ_k^{ji}(x)}.
$$
The inner sum is equivariant and belongs to the $k$th chaos. It equals $s_i^{-k}\Pi_kQ_k^{ji}$, where $Q_k^{ji}(x)=p_j((\iota_i x)^{\odot k})$ is the coordinate polynomial specified in Theorem~\ref{thm:factor}. Orthogonal projection onto chaos $k$ therefore yields \eqref{eq:factor}.

If $\sigma$ is positively homogeneous of degree one, $\sigma(sZ)=s\sigma(Z)$ for $s>0$, hence
$$
a_k(\sigma;s)=s\,a_k(\sigma;1).
$$
Every continuous degree-one positively homogeneous scalar map has the form $\sigma(x)=c_1x+c_2|x|$.

The linear term contributes only at degree one, while the even term is orthogonal to all odd Hermite polynomials. Thus every odd coefficient above degree one vanishes. For ReLU, $a_0=s/\sqrt{2\pi}$, $a_1=s/2$, and
\[
a_{2r}(\relu;s)=\frac{s\He_{2r-2}(0)}{(2r)!\sqrt{2\pi}}\ne0\quad(r\ge1).
\]
The last identity follows by two integrations by parts using $(\He_k\varphi)'=-\He_{k+1}\varphi$, where $\varphi$ is the standard normal density; $\He_{2m}(0)=(-1)^m(2m)!/(2^m m!)$. Thus ReLU has support $\{0,1\}\cup2\N_{>0}$ \citep[Theorem~4.3]{Shayestehfar2026Graded}.

\subsection{Multiplicity in the implemented feature tensor}

The theorem above is stated for a multiplicity-free permutation module. The implemented tensor is $V_{n,3}\otimes\R^6$, so each $L_i$ appears with multiplicity six. The narrow theorem is sufficient for the implemented sourcewise computation; the broader factorisation in Shayestehfar~\citep[Theorem~3.6 and Appendix~B]{Shayestehfar2026Graded} treats arbitrary inputs and multiple coordinate orbits. The group acts trivially on the channel index, the projectors act as $E_i\otimes I_6$, and coordinatewise ReLU acts independently in each channel. The factorisation is therefore applied copy by copy, with its channel gain included in the scalar activation, before invariant summaries are formed. This does not establish independent control of joint multidegree interactions between channels, nor a degree decomposition of the full trained network.

\vspace{0.3cm}
\section{Route and multiplicity calculations}
\label{app:routes}

For the triple module, the degree-$k$ route multiplicity is
$$
m_{ij}^{(k)}=\dim\HomSn(\Sym^kL_i,L_j).
$$
At $n=20$ the first nontrivial degrees are shown below. Rows are sources and columns are targets. A zero means that symmetry forbids that degree; positive entries give the multiplicity. Degree one is the identity matrix by Schur's lemma.

\begin{figure}[H]
\centering
\begin{minipage}{.31\textwidth}\centering
\textbf{$k=2$}\\[2pt]
\begin{tabular}{c|rrrr}
 & $L_0$&$L_1$&$L_2$&$L_3$\\\hline
$L_0$&1&0&0&0\\
$L_1$&1&1&1&0\\
$L_2$&1&1&2&1\\
$L_3$&1&1&2&2
\end{tabular}
\end{minipage}\hfill
\begin{minipage}{.31\textwidth}\centering
\textbf{$k=3$}\\[2pt]
\begin{tabular}{c|rrrr}
 & $L_0$&$L_1$&$L_2$&$L_3$\\\hline
$L_0$&1&0&0&0\\
$L_1$&1&2&1&1\\
$L_2$&2&3&6&5\\
$L_3$&2&5&9&16
\end{tabular}
\end{minipage}\hfill
\begin{minipage}{.31\textwidth}\centering
\textbf{$k=4$}\\[2pt]
\begin{tabular}{c|rrrr}
 & $L_0$&$L_1$&$L_2$&$L_3$\\\hline
$L_0$&1&0&0&0\\
$L_1$&2&3&3&1\\
$L_2$&5&10&22&21\\
$L_3$&12&32&84&139
\end{tabular}
\end{minipage}
\caption{Degree-resolved interaction multiplicities for $V_{20,3}$. Matrix form avoids the direction and label ambiguity of dense arrow diagrams.}
\label{fig:route-matrices}
\end{figure}

These multiplicities were computed exactly from symmetric-group characters. Writing $\chi_{V_{n,i}}$ for the permutation character on $i$-subsets gives $\chi_{L_i}=\chi_{V_{n,i}}-\chi_{V_{n,i-1}}$, and the characters of $\Sym^kL_i$ follow from Newton identities applied to the power sums $\chi_{L_i}(g^r)$. Inner products with $\chi_{L_j}$ then give $m_{ij}^{(k)}$.

\subsection{Realised coordinate maps and their norms}

A nonzero Hom space does not, in complete generality, force the particular coordinate map $Q_k^{ji}$ to be nonzero. For the benchmark family this can be checked exactly, using the projected Gram criterion of Shayestehfar~\citep[Section~3.3]{Shayestehfar2026Graded}. Let $N=\binom n3$ and $s_i^2=d_i/N$. Under $\gamma_{L_i}$ the standardised coordinate forms have correlation
$$
\rho_{bb'}=\frac{(E_i)_{bb'}}{s_i^2}.
$$
From Mehler's identity,
$\E[\He_k(Z_1)\He_k(Z_2)]=k!\rho^k$, so
$$
\|\Pi_kQ_k^{ji}\|_{L^2(\gamma_{L_i};L_j)}^2
=k!
\sum_{b,b'}(E_i)_{bb'}^k(E_j)_{bb'}.
$$
For triples at Johnson distance $t$, write $q_i(t)=N(E_i)_{bb'}$ and let
$v_t=\binom3t\binom{n-3}{t}$ be the number of $b'$ at that distance from a fixed $b$. There are $N$ choices for $b$, hence
\begin{equation}
\|\Pi_kQ_k^{ji}\|_{L^2(\gamma_{L_i};L_j)}^2
=\frac{k!}{N^k}
\sum_{t=0}^3v_t\,q_i(t)^kq_j(t).
\label{eq:mehler-corrected}
\end{equation}
The factor $N^k$ arises because the $N$ ordered choices of the first triple cancel one of the $N$ factors contributed by the $k+1$ projector entries. Since the omitted prefactor in the zero/nonzero test is strictly positive, the realised-route conclusions depend only on the finite sum in~\eqref{eq:mehler-corrected}.

Exact rational evaluation of \eqref{eq:mehler-corrected} verifies, for every $15\le n\le24$, every $i,j$ and every $k\le4$, that $Q_k^{ji}$ is nonzero exactly when the corresponding Hom space is nonzero. This is a finite verification for the benchmark family, not a general theorem. It justifies the thirteen ReLU blocks used by \SpechtTriple{} over every graph order in the experiment.

\vspace{0.3cm}
\section{Complete experimental protocol and per-seed results}
\label{app:protocol}

\subsection{Setup summary}

Summary of the experiments.

\begin{table}[H]
\centering
\caption{Experimental setup.}
\label{tab:setup}
\small
\begin{tabular}{p{0.24\textwidth}p{0.70\textwidth}}
\toprule
Item & Fixed setting \\
\midrule
Primary targets & graph diameter, algebraic connectivity \\
Graph orders & $15\le n\le24$, exact-$n$ representation in every batch \\
Families & Erd\H{o}s--R\'enyi, Barab\'asi--Albert, Watts--Strogatz, two-block SBM \\
Train/validation/test & $8{,}000/2{,}000/4{,}000$ graphs \\
Backbone & five-layer GIN, width $64$, frozen before residual training \\
Residuals & \Johnson{}, \Projector{} and \SpechtTriple{}, $884$ parameters each \\
Optimisation seeds & $11,23,37,53,71,89,107,127,149,173$ \\
Residual learning rate & $10^{-3}$ for both models \\
Score & mean of the two training-standardised test MSEs \\
\bottomrule
\end{tabular}
\end{table}

\subsection{Graph generation}

The final corpus was generated once with seed $440081$. Graph order is sampled uniformly from $15$ through $24$ and the corpus is balanced over four families. Disconnected samples are rejected, with at most $300$ attempts before a deterministic connectivity rescue that did not affect the audited final protocol.

\begin{table}[H]
\centering
\caption{Graph-generator parameters. Uniform intervals are sampled independently for each graph.}
\small
\begin{tabular}{lll}
\toprule
Family & Parameters & Range \\
\midrule
Erd\H{o}s--R\'enyi & edge probability $p$ & $p\sim U(0.08,0.34)$ \\
Barab\'asi--Albert & attachments $m$ & $m\in\{1,2,3,4\}$ \\
Watts--Strogatz & degree $k$, rewiring $\beta$ & $k\in\{2,4,6\}$, $\beta\sim U(0.05,0.45)$ \\
Two-block SBM & block cut, $p_{\mathrm{in}},p_{\mathrm{out}}$ & cut in middle third; $U(0.18,0.55)$; $U(0.02,0.18)$ \\
\bottomrule
\end{tabular}
\end{table}

The split sizes are $8{,}000/2{,}000/4{,}000$, with each family contributing $2{,}000/500/1{,}000$. Node identifiers are four-dimensional independent standard Gaussian vectors, centred and scaled coordinatewise within each graph. Targets are unweighted graph diameter and the second-smallest eigenvalue of the combinatorial Laplacian $D-A$. The training means and standard deviations are $(4.51775,1.09949)$ and $(2.33446,0.79849)$, respectively. Spectral radius is retained only as a shortcut diagnostic.

A four-round permutation-invariant $1$-WL structural hash is used for global split exclusion. Train/validation/test overlap is empty. The final corpus also excludes $11{,}981$ graph hashes inspected during earlier development, with recorded cross-corpus overlap zero.

\subsection{Architectures}

The GIN has depth $5$ and width $64$. Its input projection maps $\R^4$ to $\R^{64}$. Each layer uses an $\varepsilon$-weighted self term plus adjacency aggregation, a two-layer MLP of hidden width $128$ with SiLU, a residual connection and layer normalisation. Mean pooling gives the graph representation, followed by a $64$-hidden-unit prediction head.

All three higher-order residuals use the same triple-lift design in \eqref{eq:lift} with branch width $6$ and are added to the frozen GIN prediction. The two controls have identical parameter counts by component, so the following accounting covers all three residuals:
\begin{figure}[H]
    \centering
    \small
\begin{tabular}{p{0.42\textwidth}rr}
\toprule
Component & \SpechtTriple{} & Either control \\
\midrule
node projection & 384 & 384 \\
edge-count embedding & 24 & 24 \\
source/route or Johnson mixing & $24+156$ & $144+6$ \\
layer normalisation & 24 & 24 \\
residual head & 272 & 302 \\
\midrule
Total & \textbf{884} & \textbf{884} \\
\bottomrule
\end{tabular}
\end{figure}

Specht uses a $12\to18\to2$ head and the controls use $12\to20\to2$ heads to match total active parameter counts; no inactive padding is used.

For \SpechtTriple{}, the factor $\sqrt{48}$ in Algorithm~\ref{alg:specht} is inherited from the frozen implementation. During development, several nested structured variants were embedded in a common $48$-route reference supernet and shared this normaliser so that zeroing additional route gates preserved exact function-class inclusions. Changing the constant after freezing the model would change the evaluated checkpoint, so the original normalisation is retained and documented.

For \Johnson{}, $T_n=A_1/[3(n-3)]$ and the four basis powers are $I,T_n,T_n^2,T_n^3$. The mixed tensor is divided by $\sqrt4$ before adding the learned channel bias, then passed through ReLU. \Projector{} instead uses $\sum_iE_iXU_i$, with the same $1/\sqrt4$ rescaling and channel bias. Each control pools the mean square and mean activated value over triples, followed by layer normalisation and a $12\to20\to2$ SiLU head. Specht normalises each source before ReLU and has no corresponding pre-activation bias; these are retained differences between the evaluated designs.

\subsection{Optimisation and controlled-comparison protocol}

The ten optimisation seeds are $11,23,37,53,71,89,107,127,149,173$. The GIN is trained for at most $140$ epochs at initial learning rate $2\times10^{-3}$. Residual branches use at most $140$ epochs at $10^{-3}$. Batch size is $256$, subject to a cap of $600{,}000$ triple coordinates and an automatic memory probe. Weight decay is $10^{-5}$, gradients are clipped at $5$, validation is evaluated every four epochs, and learning rate is halved after three validation checks without improvement down to $2\times10^{-5}$. Early stopping patience is twelve checks. Automatic mixed precision is enabled. Checkpoints are selected only by validation mean standardised MSE.

The higher-order branches use the frozen GIN checkpoint for the corresponding seed, the same residual learning rate $10^{-3}$, seed IDs and training schedule. The primary comparison is \SpechtTriple{} versus \Projector{}. There is no control hyperparameter sweep in the final runner. Test predictions are aligned by graph ID, target and graph order.

The replication unit is the optimisation seed, conditional on one fixed corpus and split. Relative reductions are ratios of paired means. The reported 95\% percentile bootstrap interval uses $100{,}000$ joint resamples of the ten paired seeds (bootstrap seed $40311$). Exact paired sign-flip tests enumerate all $2^{10}$ sign assignments of seedwise MSE differences, using the mean difference as statistic; their interpretation requires sign exchangeability under the null. We report both the directional one-sided test and a two-sided sensitivity analysis. The designated primary comparison is unadjusted; taskwise and subgroup analyses are secondary.

\subsection{Per-seed results}
\begin{table}[H]
\centering
\caption{Aggregate standardised test MSE by optimisation seed.}
\small
\begin{tabular}{rrrrr}
\toprule
Seed & GIN & Control Triple & Projector Control & Specht Triple \\
\midrule
11 & 0.09307 & 0.08139 & 0.08151 & 0.06383 \\
23 & 0.09019 & 0.07012 & 0.05813 & 0.06492 \\
37 & 0.08733 & 0.07603 & 0.08714 & 0.06515 \\
53 & 0.09374 & 0.07637 & 0.06541 & 0.06288 \\
71 & 0.08861 & 0.07254 & 0.07917 & 0.06023 \\
89 & 0.09002 & 0.07533 & 0.05430 & 0.05728 \\
107 & 0.09424 & 0.07549 & 0.06422 & 0.06090 \\
127 & 0.09453 & 0.08012 & 0.07211 & 0.06220 \\
149 & 0.08920 & 0.08068 & 0.07899 & 0.06224 \\
173 & 0.09225 & 0.07590 & 0.05274 & 0.05742 \\
\midrule
Mean & 0.09132 & 0.07640 & 0.06937 & 0.06171 \\
\bottomrule
\end{tabular}
\end{table}

The primary absolute MSE reduction is $0.0076673$, the ratio-of-means reduction is $11.052\%$, the paired effect size is $d_z=0.713$, and the paired-bootstrap interval is $[2.242\%,18.272\%]$. Specht wins $7/10$ paired seeds. The exact one-sided and two-sided sign-flip $p$-values are $0.029296875$ and $0.05859375$, respectively. The taskwise reductions against \Projector{} are $14.24\%$ for diameter ($8/10$ wins) and $2.57\%$ for algebraic connectivity ($6/10$ wins). The latter remains statistically unresolved.

For context, the secondary \SpechtTriple{}--\Johnson{} comparison gives the previously reported $19.23\%$ reduction, $10/10$ wins and paired-bootstrap interval $[15.97\%,21.90\%]$. This is a different comparison from the primary $11.05\%$ result. The two controls span the same linear space at fixed $n$, but their basis parameterisations and their weights shared across $n$ differ. Neither comparison removes the normalisation and nonlinear-factorisation differences described in Section~\ref{sec:discussion}.

\subsection{Random-identifier redraw diagnostic}

The architecture is permutation invariant when node identifiers move with their vertices. It is not invariant to replacing them by independent new identifiers. On $512$ held-out graphs with $20$ redraws, the prediction standard deviation normalised by training-target scale is approximately $0.050$ for diameter and $0.077$ for algebraic connectivity. This diagnostic is why the paper describes random identifiers as nuisance stochastic probes rather than invariant graph features.

\subsection{Computational cost}
\label{app:efficiency}

We report computational measurements to quantify the cost of the
representation-resolved architecture. These measurements are descriptive
rather than part of the primary statistical comparison. In particular, the
present experiment does not claim that Specht Triple is computationally
cheaper than the generic equivariant control.

All three residual architectures have exactly \(884\) trainable parameters.
They nevertheless perform substantially different amounts of computation.
The timing table covers Control Triple and Specht Triple; it does not report a matched timing measurement for Projector Control. Control Triple applies a general \(S_n\)-equivariant Johnson-algebra
operator to the triple signal before the nonlinearity, whereas Specht Triple
first resolves the signal into irreducible source components and explicitly
evaluates multiple nonlinear source-to-target routes.

\paragraph{Benchmark protocol.}
We benchmarked the frozen models for each of the ten optimisation seeds on
the complete \(4{,}000\)-graph test set. For inference, each model received
two warm-up passes followed by nine timed passes, with CUDA synchronisation
before and after each measurement. The median of the nine repetitions was
recorded for each seed. Table~\ref{tab:efficiency} reports the mean and
standard deviation of these seed-level measurements. Peak memory denotes
peak CUDA memory allocated during the corresponding computation. Training
times and training-memory measurements are those recorded during the
original ten residual-branch training runs; the GIN backbone was frozen and
therefore has no incremental training entry.

\begin{table}[t]
\centering
\caption{Computational cost over the ten optimisation seeds. Inference time
is for the complete \(4{,}000\)-graph test set. Values are mean
\(\pm\) standard deviation across seeds. Training quantities for the two
higher-order models are incremental to the frozen GIN backbone.}
\label{tab:efficiency}
\small
\setlength{\tabcolsep}{4.5pt}
\begin{tabular}{lccc}
\toprule
 & GIN & Control Triple & Specht Triple \\
\midrule
Residual parameters
    & -- & \(884\) & \(884\) \\

Incremental training time (s)
    & -- & \(30.09\pm0.38\)
    & \(147.35\pm2.53\) \\

Peak training memory (GiB)
    & -- & \(0.167\pm0.001\)
    & \(0.679\pm0.001\) \\

Test-set inference time (s)
    & \(0.0291\pm0.0007\)
    & \(0.0508\pm0.0011\)
    & \(0.2042\pm0.0073\) \\

Latency (ms / graph)
    & \(0.00727\pm0.00017\)
    & \(0.01271\pm0.00029\)
    & \(0.05105\pm0.00183\) \\

Throughput (graphs / s)
    & \(137{,}538\pm3{,}184\)
    & \(78{,}722\pm1{,}721\)
    & \(19{,}609\pm663\) \\

Peak inference memory (GiB)
    & \(0.0162\pm0.0002\)
    & \(0.0947\)
    & \(0.5924\) \\
\bottomrule
\end{tabular}
\end{table}

The exact parameter match therefore does not imply a compute match. On
average, Specht Triple required approximately \(4.90\times\) the incremental
training time and \(4.06\times\) the peak training memory of Control Triple.
For end-to-end inference, Specht Triple required approximately
\(4.02\times\) the test-set inference time of Control Triple and
\(6.26\times\) its peak allocated inference memory.

It is also useful to separate the common GIN computation from the cost of
the higher-order residual branch. Relative to GIN alone, Control Triple
added $0.0217\pm0.0006\ {\mathrm{s}}$ to inference over the full test set, whereas Specht Triple added $0.1751\pm0.0068\ {\mathrm{s}}$. The corresponding incremental peak allocations were approximately
\(0.0785\) GiB and \(0.5762\) GiB, respectively. Thus the explicit
representation-resolved route computation is the dominant additional cost
of Specht Triple.

These measurements clarify the interpretation of the predictive results.
The reduction in test error from Control Triple to Specht Triple is not an
efficiency gain obtained by increasing parameter count: both residual
branches use \(884\) trainable parameters. Rather, Specht Triple spends
additional computation to impose a more structured nonlinear
parameterisation. These measurements describe the additional cost of the explicit route computation on the small-graph benchmark; they do not establish large-graph efficiency or isolate the reason for the predictive gain.

\paragraph{Scaling of the implementation.}
At fixed channel width, storing features on all $N=\binom n3$ triples costs $O(n^3)$ memory. The implementation applies Johnson adjacency through pair--triple incidence: if $B_{P,T}=1$ for $P\subset T$, $|P|=2$, then $A_1=B^\top B-3I$. Each triple has three incident pairs, so $B$ has $3N$ nonzeros and applying $A_1$ costs $O(Nc)$ at width $c$. The projectors are fixed-degree polynomials in $A_1$, and the route count is fixed, giving $O(n^3c)$ work and storage for the representation operations, apart from the common backbone and channel mixing. This is an operation count for the supplied sparse implementation, not a measured scaling law. Dense $N\times N$ projector storage would instead cost $O(n^6)$ and is avoided. Increasing graph order can still make the cubic feature tensor impractical; no performance or numerical-stability claim outside the tested range is made.

\vspace{0.3cm}
\section{Worked cyclic-group example}
\label{app:c8}

This example, also discussed in Shayestehfar~\citep[Section~4.3]{Shayestehfar2026Graded}, illustrates the separation between algebraically available degrees and degrees supplied by an activation without involving the graph experiment.

Let $G=C_8$ and let $H=\R C_8$ be the real regular representation with its indicator basis. Its irreducible decomposition is
$$
H\cong\mathbf1\oplus\varepsilon\oplus\rho_1\oplus\rho_2\oplus\rho_3,
$$
where $\mathbf1$ is trivial, $\varepsilon$ is the sign representation and the $\rho_s$ are two-dimensional real rotation representations. The minimal degree
$$
\min D(L,K),\qquad D(L,K)=\{k\ge0:\Hom_{C_8}(\Sym^kL,K)\ne0\},
$$
with $\min\varnothing=\infty$, is shown below, with sources on rows and targets on columns.
\begin{figure}[H]
    \centering
    \begin{tabular}{c|ccccc}
\toprule
 & $\mathbf1$ & $\varepsilon$ & $\rho_1$ & $\rho_2$ & $\rho_3$ \\
\midrule
$\mathbf1$ & 0 & $\infty$ & $\infty$ & $\infty$ & $\infty$ \\
$\varepsilon$ & 0 & 1 & $\infty$ & $\infty$ & $\infty$ \\
$\rho_1$ & 0 & 4 & 1 & 2 & 3 \\
$\rho_2$ & 0 & 2 & $\infty$ & 1 & $\infty$ \\
$\rho_3$ & 0 & 4 & 3 & 2 & 1 \\
\bottomrule
\end{tabular}
\end{figure}

For the rotation sources, this follows from the closed congruence condition
$$
\min D(\rho_s,\rho_t)=\min\{|r|:rs\equiv\pm t\pmod8\}.
$$
For example $\min D(\rho_1,\rho_2)=2$ and $\min D(\rho_1,\rho_3)=3$, whereas no integer solves $2r\equiv\pm1\pmod8$, so $\rho_2\to\rho_1$ is absent at every degree.

\begin{figure}[H]
\centering
\resizebox{0.82\textwidth}{!}{
\tikzset{
  c8node/.style={
    circle,
    draw,
    thick,
    minimum size=8mm,
    inner sep=0pt,
    font=\normalsize
  },
  c8edge/.style={
    -{Latex[length=2.2mm]}
  },
  c8loop/.style={
    min distance=4mm,
    looseness=5
  },
  c8biedge/.style={
    {Latex[length=2.2mm]}-{Latex[length=2.2mm]}
  },
  deg0/.style={
    c8edge,
    draw=blue!25,
    line width=0.6pt
  },
  deg1/.style={
    c8edge,
    draw=blue!40,
    line width=0.8pt
  },
  deg2/.style={
    c8edge,
    draw=blue!55,
    line width=1.0pt
  },
  deg3/.style={
    c8edge,
    draw=red!70,
    line width=1.4pt
  },
  deg4/.style={
    c8edge,
    draw=blue!85!black,
    line width=1.8pt
  },
  graphlabel/.style={
    font=\small
  },
  paneltitle/.style={
    font=\small\bfseries
  },
  degreelabel/.style={
    fill=white,
    circle,
    inner sep=1.5pt,
    font=\small
  }
}

\newcommand{\CeightBase}{%
  \node[c8node] (r1)   at (-2, 2.90) {$\rho_1$};
  \node[c8node] (r3)   at ( 2, 2.90) {$\rho_3$};

  \node[c8node] (r2)   at (-2, 0.45) {$\rho_2$};
  \node[c8node] (eps)  at ( 2, 0.45) {$\varepsilon$};

  \node[c8node] (triv) at ( 0.00,-2.00) {$\mathbf{1}$};
}

\newcommand{\CeightPanel}[2]{%
  \begin{tikzpicture}[scale=0.53]
    \CeightBase

    \node[paneltitle] at (0,4.1) {#1};

    \begin{pgfonlayer}{background}
      #2
    \end{pgfonlayer}
  \end{tikzpicture}%
}

\begin{minipage}{\textwidth}
\centering

\begin{tikzpicture}[scale=0.9]

  \node[c8node] (r1)   at (-2, 2.90) {$\rho_1$};
  \node[c8node] (r3)   at ( 2, 2.90) {$\rho_3$};

  \node[c8node] (r2)   at (-2, 0.45) {$\rho_2$};
  \node[c8node] (eps)  at ( 2, 0.45) {$\varepsilon$};

  \node[c8node] (triv) at ( 0.00,-2) {$\mathbf{1}$};

  \begin{pgfonlayer}{background}

    \draw[deg0] (r1) -- node[pos=0.62, degreelabel, text=blue!50!black] {$0$} (triv);
    \draw[deg0] (r2)  -- node[midway, degreelabel, text=blue!50!black] {$0$} (triv);
    \draw[deg0] (r3) -- node[pos=0.62, degreelabel, text=blue!50!black] {$0$} (triv);
    \draw[deg0] (eps) -- node[midway, degreelabel, text=blue!50!black] {$0$} (triv);

    \draw[deg0] (triv) edge[loop below, c8loop] node[midway, degreelabel, text=blue!50!black] {$0$} (triv);

    \draw[deg1] (r1)  edge[loop left, c8loop]  node[midway, degreelabel, text=blue!60!black] {$1$} (r1);
    \draw[deg1] (r2)  edge[loop left, c8loop]  node[midway, degreelabel, text=blue!60!black] {$1$} (r2);
    \draw[deg1] (r3)  edge[loop right, c8loop] node[midway, degreelabel, text=blue!60!black] {$1$} (r3);
    \draw[deg1] (eps) edge[loop right, c8loop] node[midway, degreelabel, text=blue!60!black] {$1$} (eps);

    \draw[deg2] (r1) -- node[midway, degreelabel, text=blue!70!black] {$2$} (r2);
    \draw[deg2] (r3) -- node[pos=0.3, degreelabel, text=blue!70!black] {$2$} (r2); 
    \draw[deg2] (r2) -- node[midway, degreelabel, text=blue!70!black] {$2$} (eps);

    \draw[deg3, c8biedge] (r1) -- node[midway, degreelabel, text=red!80!black] {$3$} (r3);

    \draw[deg4] (r1) -- node[pos=0.7, degreelabel, text=blue!85!black] {$4$} (eps);
    \draw[deg4] (r3) -- node[midway, degreelabel, text=blue!85!black] {$4$} (eps);

  \end{pgfonlayer}

\end{tikzpicture}

\end{minipage}
}
\caption{The graded Hom graph for $\R C_8$. Edge labels give the first degree at which the corresponding symmetric-power Hom space becomes nonzero.}
\label{fig:c8-hom}
\end{figure}
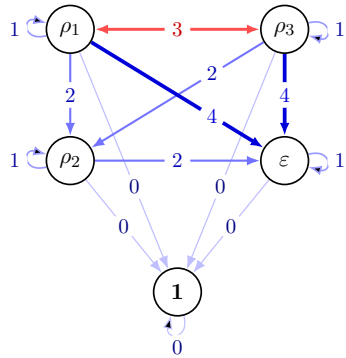

ReLU has Hermite support $\{0,1\}\cup2\N_{>0}$. The two routes $\rho_1\leftrightarrow\rho_3$ first occur in odd degree three and, in fact, only in higher odd degrees. They are therefore absent from the ReLU interaction graph even though both source and target are faithful. A translated ReLU $\sigma_c(x)=(x-c)_+$ has
$$
a_3(\sigma_c;s)=\frac{c}{6}\,\varphi(c/s),
$$
so every nonzero translation restores both cubic routes. For the regular representation of a finite abelian group, every permitted degree has a nonzero coordinate polynomial \citep[Proposition~3.14]{Shayestehfar2026Graded}. In this example it is therefore the activation coefficient that removes the otherwise realised cubic routes. This implication is special to these coordinates and is not valid for arbitrary permutation representations.

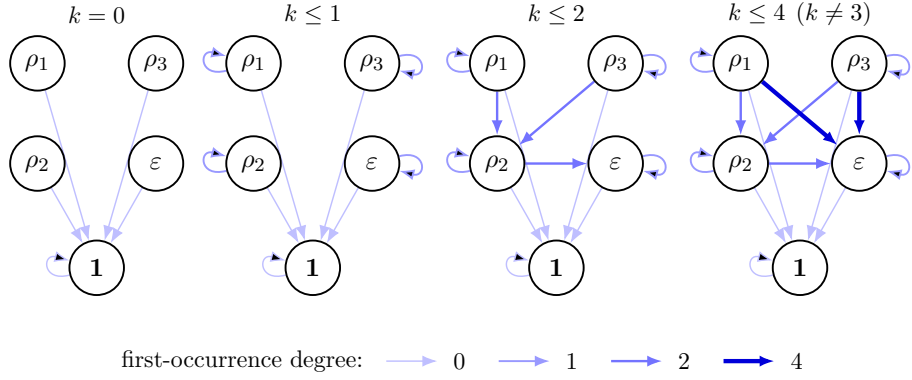
\begin{figure}[H]
\centering
\resizebox{0.92\textwidth}{!}{
\tikzset{
  c8node/.style={
    circle,
    draw,
    thick,
    minimum size=8mm,
    inner sep=0pt,
    font=\normalsize
  },
  c8edge/.style={
    -{Latex[length=2.2mm]},
    line width=0.9pt
  },
  c8loop/.style={
    min distance=4mm,
    looseness=5
  },
  deg0/.style={
    c8edge,
    draw=blue!25,
    line width=0.6pt
  },
  deg1/.style={
    c8edge,
    draw=blue!40,
    line width=0.8pt
  },
  deg2/.style={
    c8edge,
    draw=blue!55,
    line width=1.0pt
  },
  deg4/.style={
    c8edge,
    draw=blue!85!black,
    line width=1.8pt
  },
  graphlabel/.style={
    font=\small
  },
  paneltitle/.style={
    font=\small\bfseries
  }
}

\newcommand{\CeightBase}{%
  \node[c8node] (r1)   at (-1.45, 2.90) {$\rho_1$};
  \node[c8node] (r3)   at ( 1.45, 2.90) {$\rho_3$};

  \node[c8node] (r2)   at (-1.45, 0.45) {$\rho_2$};
  \node[c8node] (eps)  at ( 1.45, 0.45) {$\varepsilon$};

  \node[c8node] (triv) at ( 0.00,-2.10) {$\mathbf{1}$};
}

\newcommand{\CeightPanel}[2]{%
  \begin{tikzpicture}[scale=0.60]
    \CeightBase

    \node[paneltitle] at (0,4.1) {#1};

    \begin{pgfonlayer}{background}
      #2
    \end{pgfonlayer}
  \end{tikzpicture}%
}

\begin{minipage}{0.98\linewidth}
\centering

\begin{tabular}{
  @{}
  c
  @{\hspace{2mm}}
  c
  @{\hspace{2mm}}
  c
  @{\hspace{2mm}}
  c
  @{}
}

\CeightPanel{$k=0$}{%
  \draw[deg0] (r1)  -- (triv);
  \draw[deg0] (r2)  -- (triv);
  \draw[deg0] (r3)  -- (triv);
  \draw[deg0] (eps) -- (triv);

  \draw[deg0]
    (triv) edge[loop left,c8loop] (triv);
}
&
\CeightPanel{$k\leq 1$}{%
  \draw[deg0] (r1)  -- (triv);
  \draw[deg0] (r2)  -- (triv);
  \draw[deg0] (r3)  -- (triv);
  \draw[deg0] (eps) -- (triv);

  \draw[deg0]
    (triv) edge[loop left,c8loop] (triv);

  \draw[deg1]
    (r1) edge[loop left,c8loop] (r1);
  \draw[deg1]
    (r2) edge[loop left,c8loop] (r2);
  \draw[deg1]
    (r3) edge[loop right,c8loop] (r3);
  \draw[deg1]
    (eps) edge[loop right,c8loop] (eps);
}
&
\CeightPanel{$k\leq 2$}{%
  \draw[deg0] (r1)  -- (triv);
  \draw[deg0] (r2)  -- (triv);
  \draw[deg0] (r3)  -- (triv);
  \draw[deg0] (eps) -- (triv);

  \draw[deg0]
    (triv) edge[loop left,c8loop] (triv);

  \draw[deg1]
    (r1) edge[loop left,c8loop] (r1);
  \draw[deg1]
    (r2) edge[loop left,c8loop] (r2);
  \draw[deg1]
    (r3) edge[loop right,c8loop] (r3);
  \draw[deg1]
    (eps) edge[loop right,c8loop] (eps);

  \draw[deg2] (r1) -- (r2);
  \draw[deg2] (r3) -- (r2);
  \draw[deg2] (r2) -- (eps);
}
&
\CeightPanel{$k\leq 4 \:\:(k\neq 3)$}{%
  \draw[deg0] (r1)  -- (triv);
  \draw[deg0] (r2)  -- (triv);
  \draw[deg0] (r3)  -- (triv);
  \draw[deg0] (eps) -- (triv);

  \draw[deg0]
    (triv) edge[loop left,c8loop] (triv);

  \draw[deg1]
    (r1) edge[loop left,c8loop] (r1);
  \draw[deg1]
    (r2) edge[loop left,c8loop] (r2);
  \draw[deg1]
    (r3) edge[loop right,c8loop] (r3);
  \draw[deg1]
    (eps) edge[loop right,c8loop] (eps);

  \draw[deg2] (r1) -- (r2);
  \draw[deg2] (r3) -- (r2);
  \draw[deg2] (r2) -- (eps);

  \draw[deg4] (r1) -- (eps);
  \draw[deg4] (r3) -- (eps);
}

\end{tabular}

\par\vspace{5mm}

\begin{tikzpicture}
  \node[graphlabel,anchor=east] at (2.15,0)
    {first-occurrence degree:};

  \draw[deg0] (2.35,0) -- +(0.75,0);
  \node[graphlabel,anchor=west] at (3.20,0) {$0$};

  \draw[deg1] (4.00,0) -- +(0.75,0);
  \node[graphlabel,anchor=west] at (4.85,0) {$1$};

  \draw[deg2] (5.65,0) -- +(0.75,0);
  \node[graphlabel,anchor=west] at (6.50,0) {$2$};

  \draw[deg4] (7.30,0) -- +(0.75,0);
  \node[graphlabel,anchor=west] at (8.15,0) {$4$};
\end{tikzpicture}

\end{minipage}
}
\caption{Cumulative construction of the coordinatewise ReLU interaction graph on $\R C_8$. ReLU retains degrees $0$, $1$ and positive even degrees, so the degree-three red routes never enter.}
\label{fig:c8-relu}
\end{figure}

\end{document}